# VolumeDeform:
# Real-time Volumetric Non-rigid Reconstruction

Matthias Innmann[1]  Michael Zollhöfer[2]
Matthias Nießner[3]  Christian Theobalt[2]  Marc Stamminger[1]

[1]University of Erlangen-Nuremberg  [2]Max-Planck-Institute for Informatics  [3]Stanford University

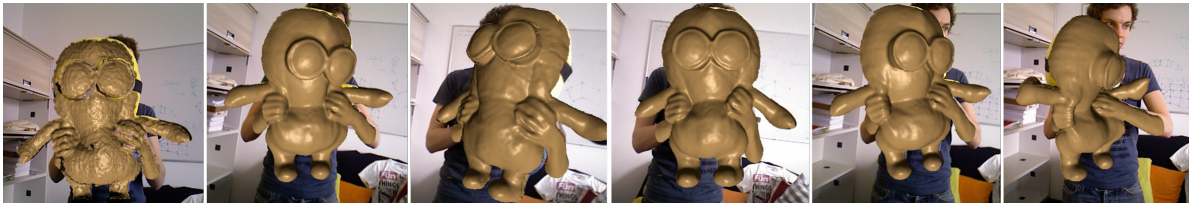

Fig. 1: Real-time non-rigid reconstruction result overlayed on top of RGB input

**Abstract.** We present a novel approach for the reconstruction of dynamic geometric shapes using a single hand-held consumer-grade RGB-D sensor at real-time rates. Our method builds up the scene model from scratch during the scanning process, thus it does not require a pre-defined shape template to start with. Geometry and motion are parameterized in a unified manner by a volumetric representation that encodes a distance field of the surface geometry as well as the non-rigid space deformation. Motion tracking is based on a set of extracted sparse color features in combination with a dense depth constraint. This enables accurate tracking and drastically reduces drift inherent to standard model-to-depth alignment. We cast finding the optimal deformation of space as a non-linear regularized variational optimization problem by enforcing local smoothness and proximity to the input constraints. The problem is tackled in real-time at the camera's capture rate using a data-parallel flip-flop optimization strategy. Our results demonstrate robust tracking even for fast motion and scenes that lack geometric features.

## 1 Introduction

Nowadays, RGB-D cameras, such as the Microsoft Kinect, Asus Xtion Pro, or Intel RealSense, have become an affordable commodity accessible to everyday users. With the introduction of these sensors, research has started to develop efficient algorithms for dense static 3D reconstruction. KinectFusion [1, 2] has shown that despite their low camera resolution and adverse noise characteristics, high-quality reconstructions can be achieved, even in real time. Follow-up work extended the underlying data structures and depth fusion algorithms in order to provide better scalability for handling larger scenes [3–6] and a higher reconstruction quality [7, 8].

While these approaches achieve impressive results on static environments, they do not reconstruct dynamic scene elements such as non-rigidly moving objects. However,

---

the reconstruction of deformable objects is central to a wide range of applications, and also the focus of this work. In the past, a variety of methods for dense deformable geometry tracking from multi-view camera systems [9] or a single RGB-D camera, even in real-time [10], were proposed. Unfortunately, all these methods require a complete static shape template of the tracked scene to start with; they then deform the template over time.

Object type specific templates limit applicability in general scenes, and are often hard to construct in practice. Therefore, template-free methods that jointly build up the shape model along with tracking its non-rigid deformations — from partial scans only — have been investigated [11–16], but none of them achieves real-time performance.

Recently, a first method has been proposed that tackles the hard joint model reconstruction and tracking problem at real-time rates: *DynamicFusion* [17] reconstructs an implicit surface representation — similar to *KinectFusion* — of the tracked object, while jointly optimizing for the scene's rigid and non-rigid motion based on a coarse warping field. Although the obtained results are impressive given the tight real-time constraint, we believe that this is not the end of the line. For instance, their depth-only model-to-frame tracking strategy cannot track tangential motion, since all color information is omitted. Without utilizing global features as anchor points, model-to-frame tracking is also prone to drift and error accumulation. In our work, we thus propose the use of sparse RGB feature matching to improve tracking robustness and to handle scenes with little geometric variation. In addition, we propose an alternative representation for the deformation warp field.

In our new algorithm, we perform non-rigid surface tracking to capture shape and deformations on a fine level of discretization instead of a coarse deformation graph. This is realized by combining as-rigid-as-possible (ARAP) volume regularization of the space embedding the surface [18] with automatically generated volumetric control lattices to abstract geometric complexity. The regular structure of the lattice allows us to define an efficient multi-resolution approach for solving the underlying non-linear optimization problem. Finally, we incorporate globally-consistent sparse SIFT feature correspondences over the complete history of observed input frames to aid the alignment process. This minimizes the risk of drift, and enables stable tracking for fast motions.

Our real-time, non-rigid volumetric reconstruction approach is grounded on the following three main contributions:

– a dense unified volumetric representation that encodes both the scene's geometry and its motion at the same resolution,
– the incorporation of global sparse SIFT correspondences into the alignment process (e.g., allowing for robust loop closures),
– and a data-parallel optimization strategy that tackles the non-rigid alignment problem at real-time rates.

## 2   Related Work

**Online Static Reconstruction**: Methods for offline static 3D shape reconstruction from partial RGB-D scans differ in the employed scene representation, such as point-based representations [19–21] or meshes [22]. In the context of commodity range sensors, implicit surface representations became popular [23–26] since they are able to efficiently regularize out noise from low-quality input data. Along with an appropriate surface representation, methods were developed that are able to reconstruct small scenes in real time [27, 28]. One prominent example for *online* static 3D scene reconstruction with a hand-held commodity sensors is *KinectFusion* [1, 2]. A dense reconstruction is obtained based on a truncated signed distance field (TSDF) [23] that is updated at framerate, and model-to-frame tracking is performed using fast variants of the Iterative Closest Point (ICP) algorithm [29]. Recently, the scene representation has been extended to scale to larger reconstruction volumes [3, 4, 30, 8, 5, 6].

**Non-rigid Deformation Tracking**: One way to handle dynamics is by tracking non-rigid surface deformations over time. For instance, objects of certain types can be non-rigidly tracked using controlled multi-RGB [31] or multi-depth [32, 33] camera input. Template-based methods for offline deformable shape tracking or performance capture of detailed deforming meshes [34–41] were also proposed. Non-rigid structure-from-motion methods can capture dense deforming geometry from monocular RGB video [42]; however, results are very coarse and reconstruction is far from real-time. The necessity to compensate for non-rigid distortions in shape reconstruction from partial RGB-D scans may also arise when static reconstruction is the goal. For instance, it is hard for humans to attain the exact same pose in multiple partial body scans. Human scanning methods address this by a non-rigid compensation of posture differences [43, 44, 11], or use template-based pose alignment to fuse information from scans in various poses [15, 45]. Real-time deformable tracking of simple motions of a wide range of objects has been demonstrated [10], but it requires a KinectFusion reconstruction of a static template before acquisition. Hence, template-free methods that simultaneously track the non-rigidly deforming geometry of a moving scene and build up a shape template over time were investigated. This hard joint reconstruction and tracking problem has mostly been looked at in an offline context [11–16, 46, 47]. In addition to runtime, drift and oversmoothing of the shape model are a significant problem that arises with longer input sequences. The recently proposed *DynamicFusion* approach [17] is the first to jointly reconstruct and track a non-rigidly deforming shape from RGB-D input in real-time (although the color channel is not used). It reconstructs an implicit surface representation - similar to the *KinectFusion* approach - while jointly optimizing for the scene's rigid and non-rigid motion based on a coarse warping field parameterized by a sparse deformation graph [48]. Our approach tackles the same setting, but uses a dense volumetric representation to embed both the reconstructed model and the deformation warp field.While *DynamicFusion* only uses geometric correspondences, we additionally employ sparse photometric feature correspondences over the complete history of frames. These features serve as global anchor points and mitigate drift, which typically appears in model-to-frame tracking methods.

## 3 Method Overview

Input to our method is a 30Hz stream captured by a commodity RGB-D sensor. At each time step $t$, a color map $\mathcal{C}_t$ and a depth map $\mathcal{D}_t$ are recorded, both at a resolution of $640 \times 480$ pixels. Color and depth are assumed to be spatially and temporally aligned. For reconstruction and non-rigid tracking of the observed scene, we use a unified volumetric representation (Sec. 4) that models both, the scene's geometry as well as its deformation. The scene is fused into a truncated signed distance field (TSDF) [23], which stores the scene's geometry and color in its initial, *undeformed* shape. A deformation field is stored at the same resolution as the TSDF in order to define a rigid transformation per voxel. In each frame, we continuously update the deformation field and fuse new RGB-D images into the undeformed shape. An overview of the steps performed each frame is shown in Fig. 2. We first generate a polygonal mesh of the shape $\mathbf{P}$, which is the current isosurface of $D$ with the current deformation field applied. Next, we search for suitable correspondences between $\mathbf{P}$ and the input depth and color map (Sec. 5), based on sparse color feature matching as well as a dense depth-based correspondence search. Based on the correspondences, we adapt the space deformation (Sec. 6) such that the scene's geometry and color best match the observed input depth and detected features. The update of the deformation field is repeated in an Iterative Closest Point (ICP) fashion. Finally, we fuse the per-frame captured depth and color data into the TSDF (Sec. 8). The underlying high-dimensional non-linear optimization problem is solved in every step using a data-parallel flip-flop iteration strategy (Sec. 7). We demonstrate online non-rigid reconstruction results at framerate and compare to template-free and template-based state-of-the-art reconstruction and tracking approaches (Sec. 9). Finally, we discuss limitations (Sec. 10) and future directions (Sec. 11).

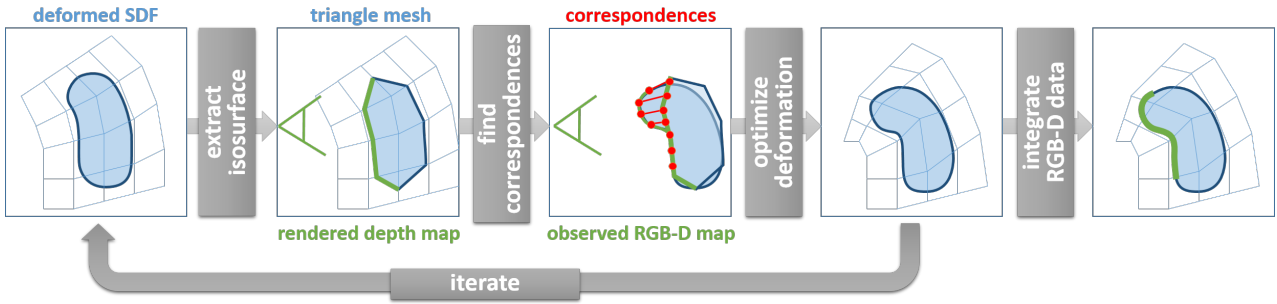

Fig. 2: Method overview: first, a deformed 3D mesh is extracted from the signed distance field using Marching Cubes. The mesh is rendered to obtain a depth map, which is used to generate dense depth correspondences. Next, we match SIFT features of the current frame with those of all previous frames. Based on all correspondences, we optimize the deformation field such that the resulting model explains the current depth and color observation. Finally, we integrate the RGB-D data of the current frame

## 4  Scene Representation

We reconstruct non-rigid scenes incrementally by joint motion tracking and surface reconstruction. The two fundamental building blocks are a truncated signed distance (TSDF) function [23] for reconstruction of the shape in its initial, *undeformed* pose and a space deformation field to track the deformations. We discretize both in a unified manner on a shared regular volumetric grid $\mathcal{G}$. The grid is composed of a set of grid points enumerated by a three-dimensional index $i$. Each grid point stores six attributes.

The first three attributes represent the scene in its undeformed pose by a truncated signed distance $D_i \in \mathbb{R}$, a color $\mathbf{C}_i \in [0, 255]^3$, and a confidence weight $W_i \in \mathbb{R}$. The zero level set of $D$ is the undeformed shape $\hat{\mathbf{P}} = D^{-1}(0)$, which we call *canonical pose* in the following. New depth data is continuously integrated into this canonical frame, where the confidence weights are used to update $D$ based on a weighted floating average (see Sec. 8). The grid points also maintain information about the current space deformation. For the $i^{th}$ gridpoint, we store its position after deformation $\mathbf{t}_i$, as well as its current local rotation $\mathbf{R}_i$, stored as three Euler angles. On top of the deformation field, we model the global motion of the scene by a global rotation $\mathbf{R}$ and translation $\mathbf{t}$. Initially, all per grid point data is set to zero, except for the positions $\mathbf{t}_i$, which are initialized to represent a regular grid. In contrast to the *DynamicFusion* approach [17], this grid-based deformation representation operates on a finer scale. Attribute values inbetween grid points are obtained via trilinear interpolation. A point $\mathbf{x}$ is deformed via the space deformation $\mathcal{S}(\mathbf{x}) = \mathbf{R} \cdot \left[ \sum_{i=1}^{|\mathcal{G}|} \alpha_i(\mathbf{x}) \cdot \mathbf{t}_i \right] + \mathbf{t}$. Here, $|\mathcal{G}|$ is the total number of grid points and the $\alpha_i(\mathbf{x})$ are the trilinear interpolation weights of $\mathbf{x}$. We denote as $\mathbf{P}$ the current deformed surface; i.e., $\mathbf{P} = \mathcal{S}(\hat{\mathbf{P}})$.

Since the deformation field stores deformation only in forward direction, an isosurface extraction via raycasting [1, 2] is not easily applicable. Thus, we use a data-parallel implementation of marching cubes [49] to obtain a polygonal representation of $\hat{\mathbf{P}}$, and then apply the deformation to the vertices. We first find all grid cells that contain a zero crossing based on a data-parallel prefix sum. One thread per valid grid cell is used to extract the final list of triangles. The resulting vertices are immediately deformed according to the current deformation field, resulting in a polygonal approximation of $\mathbf{P}$. This deformed mesh is the basis for the following correspondence association and visualization steps.

## 5 Correspondence Association

To update the deformation field, two distinct and complementary types of correspondences between the current deformed shape $\mathbf{P}$ and the new color and depth input are searched: for depth-image alignment, we perform a fast data-parallel projective lookup to obtain dense depth correspondences (see Sec. 5.1). Since in many situations depth features are not sufficient for robust tracking, we also use color information, and extract a sparse set of robust color feature correspondences (see Sec. 5.2). These also serve as global anchor points, since their descriptors are not modified over time.

### 5.1 Projective Depth Correspondences

Like most state-of-the-art online reconstruction approaches [1, 2, 17], we establish depth correspondences via a fast projective association step. Unlike them, we first extract a mesh-based representation of the isosurface $\mathbf{P}$ as described above, and then rasterize this mesh. The resulting depth buffer contains sample points $\mathbf{p}_c$ of the current isosurface. To determine a candidate for a correspondence, we project each $\mathbf{p}_c$ into the current depth map $\mathcal{D}_t$ and read the sample $\mathbf{p}_c^a$ at the target position. We generate a correspondence between $\mathbf{p}_c$ and $\mathbf{p}_c^a$, if the two points are considered sufficiently similar and appropriate for optimization. To measure similarity, we compute their world space distance $\|\mathbf{p}_c - \mathbf{p}_c^a\|_2$, and measure their normals' similarity using their dot product $\mathbf{n}_c \circ \mathbf{n}_c^a$. To make optimization more stable, we prune points close to silhouettes by looking at $\mathbf{n}_c \circ \mathbf{v}$, where $\mathbf{v}$ is the camera's view direction.

More precisely, we use three thresholds $\epsilon_d$ (distance), $\epsilon_n$ (normal deviation), and $\epsilon_v$ (view direction), and define a family of kernels $\Phi_r(x) = 1 - \frac{x}{\epsilon_r}$. If $\Phi_d(\|\mathbf{p}_c - \mathbf{p}_c^a\|_2) < 0$, $\Phi_n(1 - \mathbf{n}_c \circ \mathbf{n}_c^a) < 0$ or $\Phi_v(1 - \mathbf{n}_c \circ \mathbf{v}) < 0$, the correspondence is pruned by setting the confidence weight associated with the correspondence to zero $w_c = 0$. For valid correspondences the confidence is $w_c = \left(\frac{\Phi_d(\|\mathbf{p}_c - \mathbf{p}_c^a\|_2) + \Phi_n(1 - \mathbf{n}_c \circ \mathbf{n}_c^a) + \Phi_v(1 - \mathbf{n}_c \circ \mathbf{v})}{3}\right)^2$.

## 5.2 Robust Sparse Color Correspondences

We use a combination of dense and sparse correspondences to improve stability and reduce drift. To this end, we compute SIFT [50, 51] matches to all previous input frames on the GPU. Feature points are lifted to 3D and stored in the canonical pose by applying $\mathcal{S}^{-1}$ after detection. When a new frame is captured, we use the deformation field to map all feature points to the previous frame. We assume a rigid transform for the matching between the previous and the current frame. The rest of the pipeline is split into four main components: keypoint detection, feature extraction, correspondence association, and correspondence pruning.

**Keypoint Detection**: We detect keypoint locations as scale space maxima in a DoG pyramid of the grayscale image using a data-parallel feature detection approach. We use 4 octaves, each with 3 levels. Only extrema with a valid associated depth are used, since we later lift the keypoints to 3D. All keypoints on the same scale are stored in an array. Memory is managed via atomic counters. We use at most 150 keypoints per image. For rotational invariance, we associate each keypoint with up to 2 dominant gradient orientations.

**Feature Extraction**: We compute a 128-dimensional SIFT descriptor for each valid keypoint. Each keypoint is thus composed of its 3D position, scale, orientation, and SIFT descriptor. Our GPU implementation extracts keypoints and descriptors in about 6ms at an image resolution of $640 \times 480$.

**Correspondence Association**: Extracted features are matched with features from all previous frames using a data-parallel approach (all extracted features are stored for matching in subsequent frames). We exhaustively compute all pairwise feature distances from the current to all previous frames and vice versa. The best matching features in both directions are determined by minimum reductions in shared memory. We use at most 128 correspondences between two frames.

**Correspondence Pruning**: Correspondences are sorted based on feature distance using shared memory bubble sort. We keep the 64 best correspondences per image pair. Correspondences with keypoints not close enough in feature space, screen space, or 3D space are pruned.

## 6  Deformation Energy

To reconstruct non-rigid surfaces in real time, we have to update the space deformation $\mathcal{S}$ at sensor rate. We estimate the corresponding global pose parameters using dense projective ICP [29].

For simplicity of notation, we stack all unknowns of local deformations in a single vector:
$$\mathbf{X} = (\underbrace{\cdots, \mathbf{t}_i^T, \cdots}_{3|\mathcal{G}|\ coordinates} \ | \ \underbrace{\cdots, \mathbf{R}_i^T, \cdots}_{3|\mathcal{G}|\ angles})^T \ .$$

To achieve real-time performance, even for high-resolution grids, we cast finding the best parameters as a non-linear variational optimization problem. Based on these definitions, we define the following highly non-linear registration objective:

$$E_{total}(\mathbf{X}) = \underbrace{w_s E_{sparse}(\mathbf{X}) + w_d E_{dense}(\mathbf{X})}_{data\ term} + \underbrace{w_r E_{reg}(\mathbf{X})}_{prior\ term} \ . \quad (1)$$

The objective is composed of two data terms that enforce proximity to the current input, and a prior for regularization. The prior $E_{reg}$ regularizes the problem by favoring smooth and locally rigid deformations. The data terms are a sparse feature-based alignment objective $E_{sparse}$ and a dense depth-based correspondence measure $E_{dense}$. The weights $w_s$, $w_d$, and $w_r$ control the relative influence of the different objectives and remain constant for all shown experiments. In the following, we explain the different terms of our energy in more detail.

**Point-to-Plane Alignment**: We enforce dense alignment of the current surface $\mathbf{P}$ with the captured depth data based on a point-to-plane distance metric. The point-to-plane metric can be considered a first order approximation of the real surface geometry. This allows for sliding, which is especially useful given translational object or camera motion. To this end, we first extract a triangulation using marching cubes and rasterize the resulting mesh to obtain sample point $\mathcal{S}(\hat{\mathbf{p}}_c)$ on the isosurface. Target positions $\mathbf{p}_c^a$ are computed based on our projective correspondence association strategy presented in the previous section. The objective is based on the extracted $C$ correspondences:

$$E_{dense}(\mathbf{X}) = \sum_{c=1}^{C} w_c \cdot \left[ \left( \mathcal{S}(\hat{\mathbf{p}}_c) - \mathbf{p}_c^a \right)^T \cdot \mathbf{n}_c^a \right]^2 \ . \quad (2)$$

Here, $\mathbf{n}_c^a$ is the normal vector at $\mathbf{p}_c^a$ and $w_c$ denotes the confidence of the correspondence, see previous section.

**Sparse Feature Alignment**: In addition to the dense depth correspondence association, we use the set of $S$ sparse color-based SIFT matches (see Sec. 5) as constraints in the optimization. Let $\hat{\mathbf{f}}_s$ be the position of the $s^{th}$ SIFT feature match in the canonical frame and $\mathbf{f}_s$ its current world space position. Sparse feature alignment is enforced by:

$$E_{sparse}(\mathbf{X}) = \sum_{s=1}^{S} \|\mathcal{S}(\hat{\mathbf{f}}_s) - \mathbf{f}_s\|_2^2 \ . \quad (3)$$

This term adds robustness against temporal drift and allows to track fast motions.

**Prior Term**: Since we operate on a fine volumetric grid, rather than a coarse deformation graph, we need an efficient regularization strategy to make the highly underconstrained non-rigid tracking problems well posed. To this end, we impose the as-rigid-as-possible (ARAP) [18] prior on the grid:

$$E_{reg}(\mathbf{X}) = \sum_{i \in \mathcal{M}} \sum_{j \in \mathcal{N}_i} \left\| (\mathbf{t}_i - \mathbf{t}_j) - \mathbf{R}_i(\hat{\mathbf{t}}_i - \hat{\mathbf{t}}_j) \right\|_2^2 . \quad (4)$$

Here, $\mathcal{N}_i$ is the one-ring neighborhood of the $i^{th}$ grid point and $\mathcal{M}$ is the set of all grid points used during optimization. In our approach, $\mathcal{M}$ is the isosurface plus its one-ring. This prior is highly non-linear due to the rotations $\mathbf{R}_i$. It measures the residual non-rigid component of the deformation, which we seek to minimize.

## 7   Parallel Energy Optimization

Finding the optimum $\mathbf{X}^*$ of the tracking energy $E_{total}$ is a high-dimensional non-linear least squares problem in the unknown parameters. In fact, we only optimize the values in a one-ring neighborhood $\mathcal{M}$ around the isosurface. The objective thus has a total of $6N$ unknowns (3 for position and 3 for rotation), with $N = |\mathcal{M}|$. For the minimization of this high-dimensional non-linear objective at real-time rates, we propose a novel hierarchical data-parallel optimization strategy. First, we describe our approach for a single hierarchy level.

### 7.1   Per-Level Optimization Strategy

Fortunately, the non-linear optimization objective $E_{total}$ can be split into two independent subproblems [18] by employing an iterative flip-flop optimization strategy: first, the rotations $\mathbf{R}_i$ are fixed and we optimize for the best positions $\mathbf{t}_i$. Second, the positions $\mathbf{t}_i$ are considered constant and the rotations $\mathbf{R}_i$ are updated. These two step are iterated until convergence. The two resulting subproblems can both be solved in a highly efficient data-parallel manner, as discussed in the following.

**Data-Parallel Rotation Update**: Solving for the best rotations is still a non-linear optimization problem. Fortunately, this subproblem is equivalent to the shape matching problem [52] and has a closed-form solution. We obtain the best fitting rotation based on Procrustes analysis [53, 54] with respect to the canonical pose. Since the per grid point rotations are independent, we solve for all optimal rotations in parallel. To this end, we run one thread per gridpoint, compute the corresponding cross-covariance matrix and compute the best rotation based on SVD. With our data-parallel implementation, we can compute the best rotations for 400K voxels in 1.9ms.

**Data-Parallel Position Update**: The tracking objective $E_{total}$ is a quadratic optimization problem in the optimal positions $\mathbf{t}_i$. We find the optimal positions by setting the corresponding partial derivatives $\frac{\partial E_{total}(\mathbf{X})}{\partial \mathbf{t}_i} = \mathbf{0}$ to zero, which yields $(\mathbf{L} + \mathbf{B}^T \mathbf{B}) \cdot \mathbf{t} = \mathbf{b}$ .

Here, $\mathbf{L}$ is the Laplacian matrix, $\mathbf{B}$ encodes the point-point and point-plane constraints (including the tri-linear interpolation of positions). The right-hand side $\mathbf{b}$ encodes the fixed rotations and the target points of the constraints. We solve the linear system of equations using a data-parallel preconditioned conjugate gradient (PCG) solver, similar to [55, 10, 56–58], which we run on the GPU. Since the matrix $\mathbf{L}$ is sparse, we compute it on-the-fly in each iteration step. In contrast, $\mathbf{B}^T\mathbf{B}$ has many non-zero entries, due to the involved tri-linear interpolation. In addition, each entry is computationally expensive to compute, since we have to sum per-voxel over all contained constraints. This is a problem, especially on the coarser levels of the hierarchy, since each voxel may contain several thousand correspondences. To alleviate this problem, we pre-compute and cache $\mathbf{B}^T\mathbf{B}$, before the PCG iteration commences. In every PCG step, we read the cached values which remain constant across iterations.

### 7.2 Hierarchical Optimization Strategy

This efficient flip-flop solver has nice convergence properties on coarse resolution grids, since updates are propagated globally within only a few steps. On finer resolutions, which are important for accurate tracking, spatial propagation of updates would require too many iterations. This is a well known drawback of iterative approaches, which are known to deal well with high-frequency errors, while low-frequency components are only slowly resolved. To alleviate this problem, we opt for a *nested* coarse-to-fine optimization strategy. This provides a good trade-off between global convergence and runtime efficiency. We solve in a coarse-to-fine fashion and prolongate the solutions to the next finer level to jump-start the optimization. When downsampling constraints, we gather all constraints of a parent voxel from its 8 children on the next finer level. We keep all constraints on coarser levels and express them as a tri-linear combination of the coarse grid points.

## 8 Fusion

The depth data $\mathcal{D}_t$ of each recorded RGB-D frame is incrementally fused into the canonical TSDF following the non-rigid fusion technique introduced in DynamicFusion [17]. Non-rigid fusion is a generalization of the projective truncated signed distance function integration approach introduced by [23]. [17] define the warp field through the entire canonical frame. In contrast, we only integrate into voxels of $\mathcal{M}$ (one-ring of the current isosurface) that have been included in the optimization for at least $K_{min} = 3$ optimization steps. This ensures that data is only fused into regions with well-defined space deformations; otherwise, surface geometry may be duplicated. During runtime, the isosurface is expanding to account for previously unseen geometry. This expansion also adds new points to the grid to account for voxels that become for the first time part of $\mathcal{M}$. The position and rotation attributes of these grid points do not match the current space deformation, since they have not yet been included in the optimization. Therefore, we initialize the position $\mathbf{t}_i$ and rotation $\mathbf{R}_i$ of each new grid point by extrapolating the current deformation field. This jump-starts the optimization for the added variables.

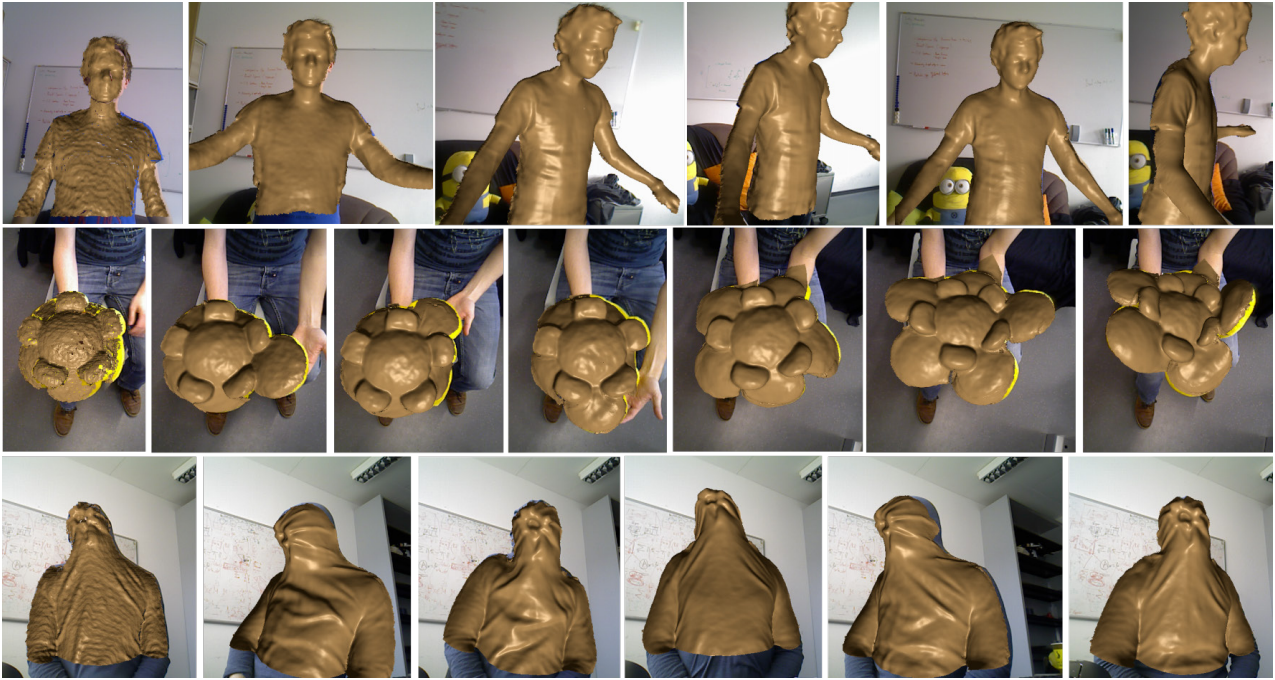

Fig. 3: A variety of non-rigid scenes reconstructed with our approach at real-time rates: UPPER BODY, SUNFLOWER, and HOODIE (top to bottom)

## 9 Results

We demonstrate a variety of non-rigid reconstruction results in Fig. 1 and Fig. 3. For a list of parameter values and additional results, we refer to the supplemental material and the accompanying video. Runtime performance and convergence analysis of our solver is also provided in the supplemental document.
In all examples, we capture an RGB-D stream using an Asus Xtion PRO, a KinectV1-style range sensor. We would like to point out that all reconstructions are obtained in real-time using a commodity desktop PC (timings are provided in the supplemental material). In addition, our method does not require any pre-computation, and we do not rely on a pre-scanned template model – all reconstructions are built from scratch.

**Importance of Sparse Color Correspondences**: A core aspect of our method is the use of sparse RGB features as global anchor points for robust tracking. Fig. 4 illustrates the improvement achieved by including the SIFT feature alignment objective. If the input lacks geometric features, dense depth-based alignment is ill-posed and results in drift, especially for tangential motion. By including color features, we are able to successfully track and reconstruct these cases.

**Comparison to Template-based Approaches**: In Fig. 5, we compare against the template-tracking method of Li et al. [39], which runs offline. Since their method uses a high-quality pre-scanned template model obtained from a static reconstruction, we can quantitatively evaluate the reconstruction generated from the dynamic sequence. To this end, we compute the geometric distance of our final reconstruction (canonical pose) to the template mesh of the first frame; see Fig. 5, right. The average error in non-occluded regions is 1mm; occluded regions cannot be reconstructed.

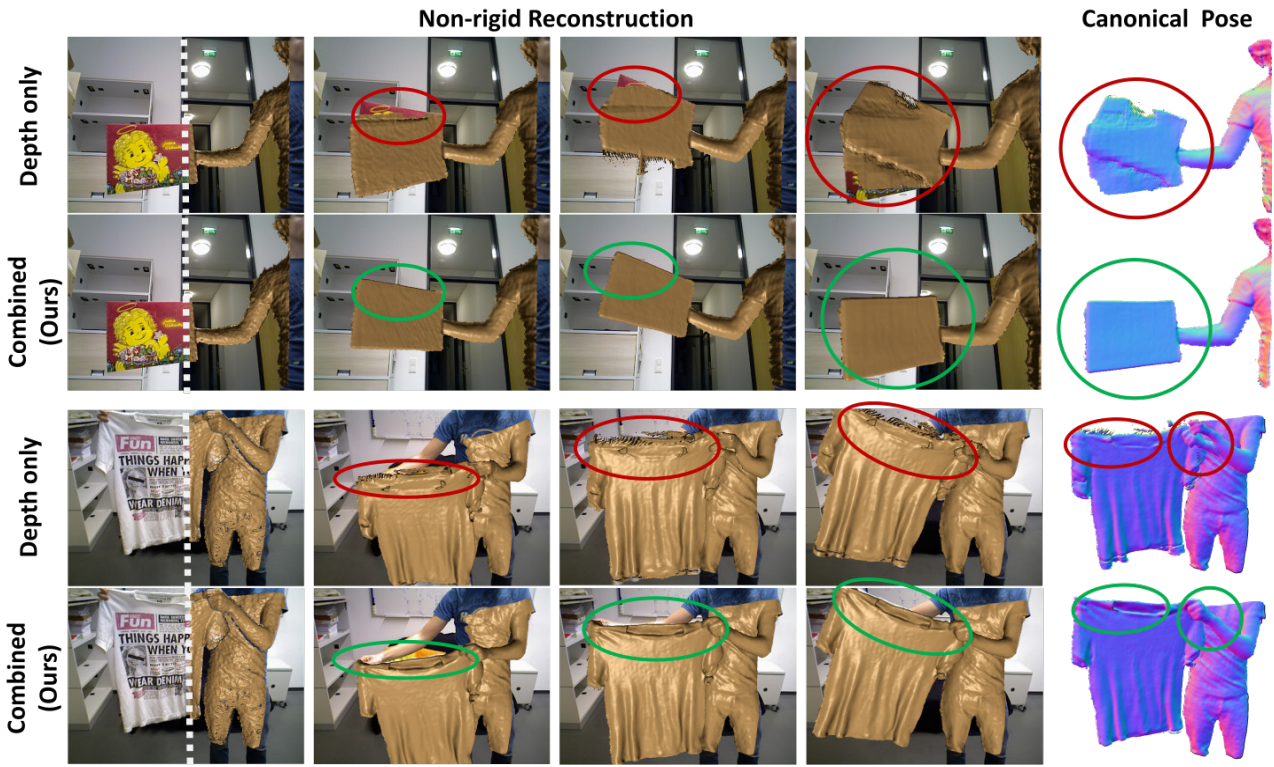

Fig. 4: Comparison of reconstructions with and without our sparse color alignment objective. Whereas depth-only reconstruction fails for tangential motion and objects with few geometric features, we achieve robust reconstructions using color features

We further compare our approach to the real-time template-tracking method by Zollhöfer et al. [10]; see Fig. 6. Even though our 3D model is obtained on-the-fly, the reconstruction quality is similar, or even higher.

**Comparison to Template-free Approaches**: Currently, *DynamicFusion* [17] is the only non-rigid reconstruction method that runs online and does not require a pre-scanned template. In Fig. 7, we compare our approach against DynamicFusion on two scenes used in their publication. Overall, we obtain at least comparable or even higher quality reconstructions. In particular, our canonical pose is of higher quality – we attribute this to the key differences in our method: first, our sparse RGB feature term mitigates drift and makes tracking much more robust (for the comparison /w and w/o SIFT feature matching, see Fig. 4). Second, our deformation field is at a higher resolution level than the coarse deformation proxy employed in DynamicFusion. This enables the alignment of fine-scale deformations and preserves detail in the reconstruction (otherwise newly-integrated frames would smooth out detail). Unfortunately, a quantitative evaluation against DynamicFusion is challenging, since their method is hard to reproduce (their code is not publicly available and not all implementation details are given in the paper).

**Stability of our Tracking**: In Fig. 8, we demonstrate the tracking stability of our method with a simple visualization: we color every surface point according to its position in the canonical grid. In the case of successful non-rigid tracking, surface color remains constant; in case of tracking failure or drift, the surface would change its color over time.

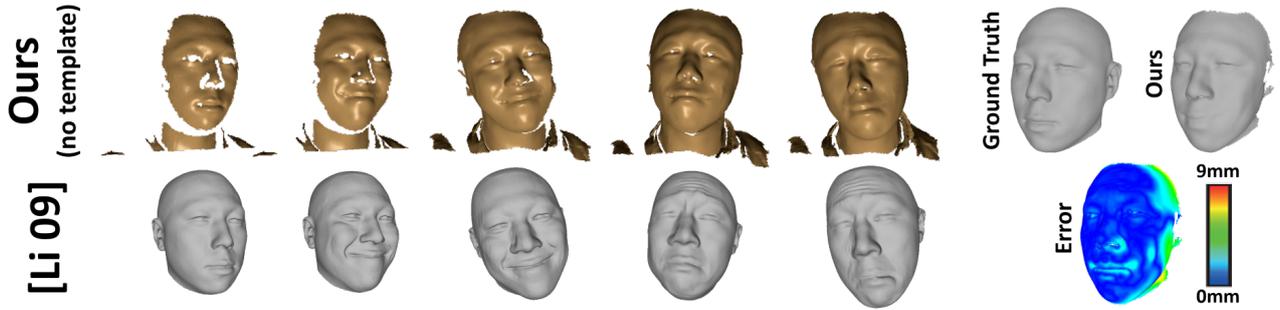

Fig. 5: Comparison to the template-based approach of Li et al. [39]: we obtain similar quality reconstructions without requiring an initial template model. On the right, we quantitatively evaluate the reconstruction quality: we compute the geometric distance of our final reconstruction (canonical pose) to the template mesh of the first frame, which is obtained from a high-quality, static pre-scanned reconstruction

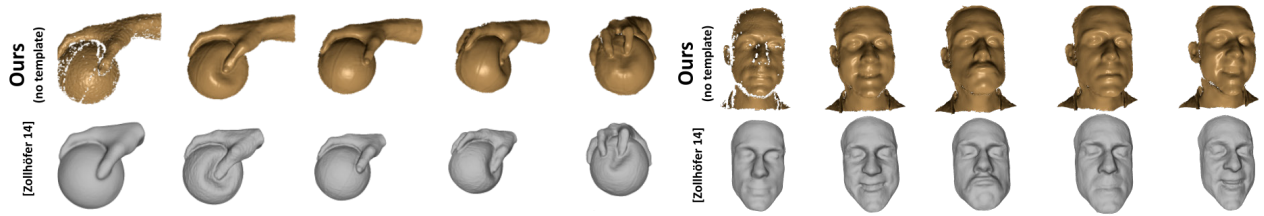

Fig. 6: Comparison to the template-based approach of Zollhöfer et al. [10]: although our reconstruction is from scratch and does not require an initial template model, we obtain reconstructions of similar quality

As we can see, our method is able to track surface points faithfully throughout the entire sequence, and all points remain stable at their undeformed positions; i.e., no drift occurs.

In Fig. 9, we evaluate the tracking stability regarding fast motions and homogeneous textures. We reconstruct the SUNFLOWER scene by only using every $n^{th}$ input frame ($n = 2, \ldots, 6$). This simulates motion of $2\times - 6\times$ speed. As can be seen, tracking remains stable up to $\approx 3\times$ speed. For higher speedups, tracking failures occur, thus leading to reconstruction errors.

**Importance of Grid Resolution and Combined Dense and Sparse Tracking**: We evaluate the importance of the fine warp-field resolution as well as the relevance of our sparse color feature term in terms of obtained deformation quality; see Fig. 10. For a low-resolution deformation grid, the warp field is not flexible enough and fine-scale deformations cannot be handled. If we use only depth data, tracking is considerably less accurate leading to local drift and may even fail completely if no geometric features are present. Only for high-resolution deformation grids and our combined tracker, drift is reduced and the texture can be reconstructed at a good quality. Note that our grids have a significantly higher number of degrees of freedom than the coarse deformation graph employed by DynamicFusion [17]; in their examples, they use only about 400 deformation nodes. We can only speculate, but based on their low-resolution warp field, DynamicFusion cannot reconstruct RGB textures.

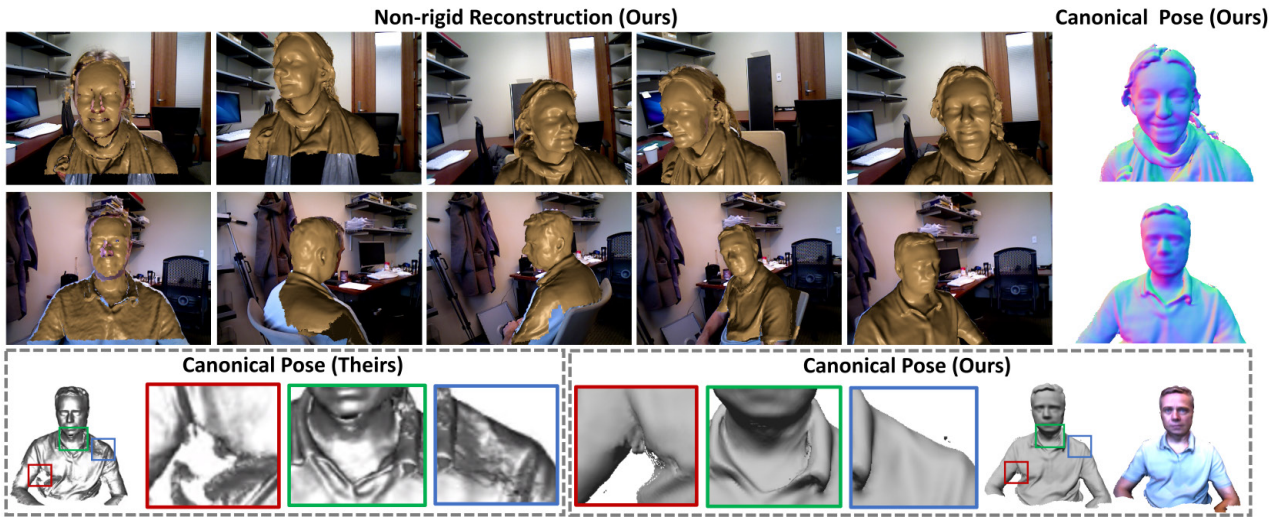

Fig. 7: Comparison to *DynamicFusion* [17]: we obtain at least comparable or even higher quality reconstructions. In particular, our canonical pose is of higher quality, since our warp field has a higher resolution than a coarse deformation proxy. In addition, our sparse feature alignment objective mitigates drift and enables more robust tracking

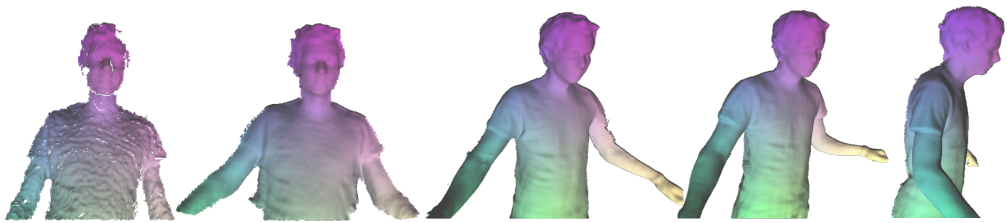

Fig. 8: Evaluation of tracking stability: surface points are colored according to the position in the canonical pose. Our non-rigid tracking maps each surface point close to its undeformed position. In case of tracking failures or drift, the surface would change its color over time

## 10   Limitations

While we are able to demonstrate compelling results and our method works well on a variety of examples, there are still limitations. First of all, robust tracking is fundamentally hard in the case of non-rigid deforming surfaces. Although global SIFT matching helps to improve robustness and minimizes alignment errors, drift is not completely eliminated. Ideally, we would like to solve a non-rigid, global bundle adjustment problem, which unfortunately exceeds the real-time computational budget.

High levels of deformation, such as fully bending a human arm, may cause problems, as our regularizer distributes deformations smoothly over the grid. We believe that adaptive strategies will be a key in addressing this issue; e.g., locally adjusting the rigidity.

Another limitation is the relatively small spatial extent that can be modeled with a uniform grid. We believe a next step on this end would be the combination of our method with a sparse surface reconstruction approach; e.g., [5, 6]. Nonetheless, we believe that our method helps to further improve the field of non-rigid 3D surface reconstruction, which is both a fundamentally hard and important problem.

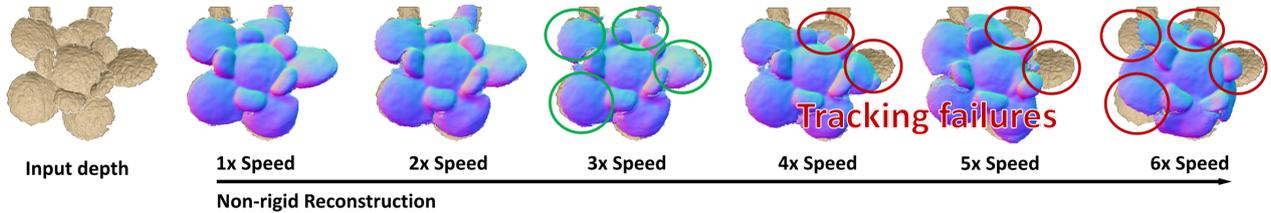

Fig. 9: Temporal Coherence: we skip every $n^{th}$ frame of the SUNFLOWER sequence. Tracking remains stable up to a $3\times$ speedup. Beyond this, tracking quality degrades.

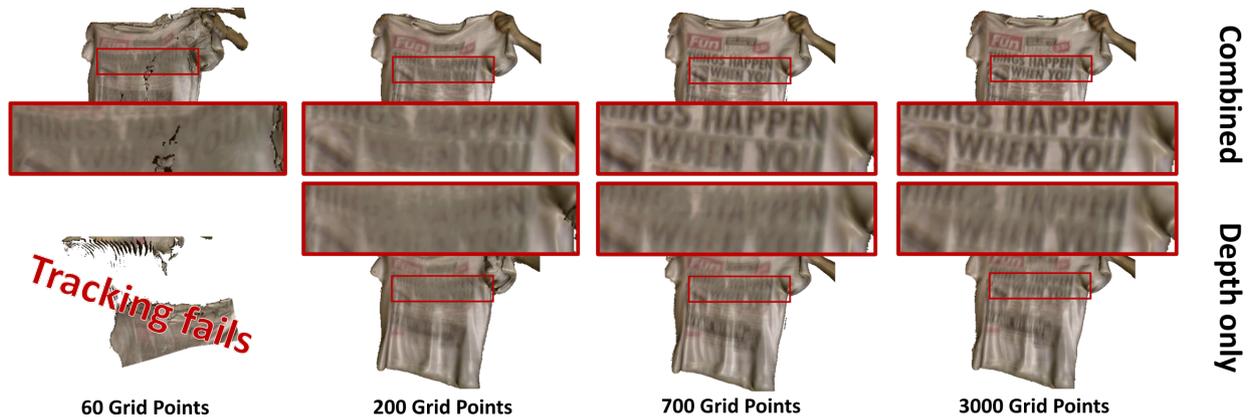

Fig. 10: Impact of grid resolution and color features: low-resolution warp fields (left) cannot capture fine-scale deformations leading to drift and blur. Depth-only tracking (bottom) also results in drift and blur. In contrast, our combined approach together with a high-resolution grid (top right) mitigates drift and leads to sharp textures

## 11  Conclusion

We present a novel approach to jointly reconstruct the geometric shape as well as motion of an arbitrary non-rigidly deforming scene at real-time rates. The foundation is a novel unified volumetric representation that encodes both, geometry and motion. Motion tracking uses sparse color as well as dense depth constraints and is based on a fast GPU-based variational optimization strategy. Our results demonstrate non-rigid reconstruction results, even for scenes that lack geometric features. We hope that our method is another stepping stone for future work, and we believe that it paves the way for new applications in VR and AR, where the interaction with arbitrary non-rigidly deforming objects is of paramount importance.

## Acknowledgments

We thank Angela Dai for the video voice over and Richard Newcombe for the DynamicFusion comparison sequences. This research is funded by the German Research Foundation (DFG) – grant GRK-1773 Heterogeneous Image System –, the ERC Starting Grant 335545 CapReal, the Max Planck Center for Visual Computing and Communications (MPC-VCC), and the Bayerische Forschungsstiftung (For3D).